\documentclass{article} 
\usepackage{nips15submit_e,times}
\usepackage{hyperref,natbib,graphicx}
\usepackage{url,amsmath,amsfonts,multirow,booktabs,algorithmic,algorithm}

\title{Empirical Gaussian priors for cross-lingual transfer learning}

\author{
Anders S{\o}gaard \\
Center for Language Technology\\
University of Copenhagen\\
Njalsgade 140, DK-2300 Copenhagen \\
\texttt{soegaard@hum.ku.dk} 
}

\nipsfinalcopy 

\begin{document}

\maketitle

\begin{abstract}
Sequence model learning algorithms typically maximize log-likelihood minus the norm of the model (or minimize Hamming loss + norm). In cross-lingual part-of-speech (POS) tagging, our target language training data consists of sequences of sentences with word-by-word labels projected from translations in $k$ languages for which we have labeled data, via word alignments. Our training data is therefore very noisy, and if Rademacher complexity is high, learning algorithms are prone to overfit. Norm-based regularization assumes a constant width and zero mean prior. We instead propose to use the $k$ source language models to estimate the parameters of a Gaussian prior for learning new POS taggers. This leads to significantly better performance in multi-source transfer set-ups. We also present a drop-out version that injects (empirical) Gaussian noise during online learning. Finally, we note that using empirical Gaussian priors leads to much lower Rademacher complexity, and is superior to optimally weighted model interpolation. 
\end{abstract}

\section{Cross-lingual transfer learning of sequence models}

The people of the world speak about 6,900 different languages. Open-source off-the-shelf natural language processing (NLP) toolboxes like OpenNLP\footnote{\url{https://opennlp.apache.org/}} and CoreNLP\footnote{\url{http://nlp.stanford.edu/software/corenlp.shtml}} cover only 6--7 languages, and we have sufficient labeled training data for inducing models for about 20--30 languages. In other words, supervised sequence learning algorithms are not sufficient to induce POS models for but a small minority of the world's languages. 

What can we do for all the languages for which no training data is available? Unsupervised POS induction algorithms have methodological problems (in-sample evaluation, community-wide hyper-parameter tuning, etc.), and performance is prohibitive of downstream applications. Some work on unsupervised POS tagging has assumed other resources such as tag dictionaries \citep{Li:ea:12}, but such resources are also only available for a limited number of languages. In our experiments, we assume that {\em no} training data or tag dictionaries are available. Our only assumption is a bit of text translated into multiple languages, specifically, fragments of the Bible. We will use Bible data for annotation projection, as well as for learning cross-lingual word embeddings (\S 3). 

Unsupervised learning with typologically informed priors \citep{Naseem:ea:10} is an interesting approach to unsupervised POS induction that is more applicable to low-resource languages. Our work is related to this work, but we learn informed priors rather than stipulate them and combine these priors with {\em annotation projection} (learning from noisy labels) rather than unsupervised learning. 

Annotation projection refers to transferring annotation from one or more source languages to the target language (for which no labeled data is otherwise available), typically through word alignments. In our experiments below, we use an unsupervised word alignment algorithm to align $15\times 12$ language pairs. For 15 languages, we have predicted POS tags for each word in our multi-parallel corpus. For each word in one of our 12 target language training datasets, we thus have up to 15 votes for each word token, possibly weighted by the confidence of the word alignment algorithm. In this paper, we simply use the majority votes. This is the set-up assumed throughout in this paper (see \S3 for more details): 

\paragraph{Low-resource cross-lingual POS tagging} We have at our disposal $k$ (=15) source language models and a multi-parallel corpus (the Bible) that we can use to project annotation from the $k$ source languages to new target languages for which no labeled data is available. If we use $k>1$ source languages, we refer to this as {\it multi-source}~cross-lingual transfer; if we only use a single source language, we refer to this as {\it single-source}~cross-lingual transfer. In this paper, we only consider multi-source cross-language transfer learning. 

Since the training data sets for our target languages (the annotation projections) are very noisy, the risk of over-fitting is extremely high. We are therefore interested in learning algorithms that efficiently limit the Rademacher complexity of the learning problem, i.e., the chance of fitting to random noise. In other words, we want a model with higher integrated bias and lower integrated variance \citep{Geman:ea:92}. Our approach -- using empirical Gaussian priors -- is introduced in \S 2, including a drop-out version of the regularizer. \S3 describes our experiments. In \S4, we provide some observations, namely that using empirical Gaussian priors reduces (i) Rademacher complexity and (ii) integrated variance, and (iii) that using empirical Gaussian priors is superior to optimally weighted model interpolation. 

\section{Empirical Gaussian priors}

We will apply empirical Gaussian priors to linear-chain conditional random fields (CRFs; \cite{Lafferty:ea:01}) and averaged structured perceptrons \citep{Collins:02}. Linear-chain CRFs are trained by maximising the conditional log-likelihood of labeled sequences $LL(\mathbf{w},\mathcal{D})=\sum_{\langle\mathbf{x},\mathbf{y}\rangle\in\mathcal{D}}\log P(\mathbf{y}|\mathbf{x})$ with $\mathbf{w}\in\mathbb{R}^m$ and $\mathcal{D}$ a dataset consisting of sequences of discrete input symbols $\mathbf{x}=x_1,\ldots,x_n$ associated with sequences of discrete labels $\mathbf{y}=y_1,\ldots,y_n$.  L$k$-regularized CRFs maximize $LL(\mathbf{w},\mathcal{D})-|\mathbf{w}|^k$ with typically $k\in\{0,1,2,\infty\}$, which all introduce costant-width, zero-mean regularizers. We refer to L$k$-regularized CRFs as {\sc L2-CRF}. L$k$ regularizers are parametric priors where the only parameter is the width of the bounding shape. The L2-regularizer is a Gaussian prior with zero mean, for example. The regularised log-likelihood with a Gaussian prior is $LL(\mathbf{w},\mathcal{D})-\frac{1}{2}\sum^m_j\left(\frac{\lambda_j-\mu_j}{\sigma^2_j}\right)^2$. For practical reasons, hyper-parameters $\mu_j$ and $\sigma_j$ are typically assumed to be constant for all values of $j$. This also holds for recent work on parametric noise injection, e.g., \citet{Soegaard:13:naacl}. If these parameters are assumed to be constant, the above objective becomes equivalent to L2-regularization. However, you can also try to learn these parameters. In empirical Bayes \citep{Casella:85}, the parameters are learned from $\mathcal{D}$ itself. \citet{Smith:Osborne:05} suggest learning the parameters from a validation set. In our set-up, we do not assume that we can learn the priors from training data (which is noisy) or validation data (which is generally not available in cross-lingual learning scenarios). Instead we estimate these parameters directly from source language models. 

When we estimate Gaussian priors from source language models, we will learn which features are invariant across languages, and which are not. We thereby introduce an ellipsoid regularizer whose centre is the average source model. In our experiments, we consider {\em both}~the case where variance is assumed to be constant -- which we call L2-regularization with priors ({\sc L2-Prior}) --- and the case where both variances and means are learned -- which we call empirical Gaussian priors ({\sc EmpGauss}). {\sc L2-Prior} is the {\sc L2-CRF} objective with $\sigma^2_j=C$ with $C$ a regularization parameter, and $\mu_j=\hat{\mu_j}$ the average value of the corresponding parameter in the observed source models. {\sc EmpGauss} replaces the above objective with $LL(\lambda)+\sum_j\log \frac{1}{\sigma\sqrt{2\pi}}e^{-\frac{(\lambda_j-\mu_j)^2}{2\sigma^2}}$, which, assuming model parameters are mutually independent, is the same as jointly optimising model probability and likelihood of the data. Note that minimizing the squared weights is equivalent to maximizing the log
probability of the weights under a zero-mean Gaussian prior, and in the same way, this is equivalent to minimising the above objective with empirically estimated parameters $\hat{\mu_j}$ and $\sigma{\mu_j}$. In other words, empirical Gaussian priors are bounding ellipsoids on the hypothesis space with learned widths and centres. Also, note that in single-source cross-lingual transfer learning, observed variance is zero, and we therefore replace this with a regularization parameter $C$ shared with the baseline. In the single-source set-up, {\sc L2-Prior} is thus equivalent to {\sc EmpGauss}. We use L-BFGS to maximize our baseline L2-regularized objectives, as well as our empirical Gaussian prior objectives. 

\paragraph{Practical observations} (i) Using empirical Gaussian priors does not assume identical feature representations in the source and target models. Model parameters for which features were unseen in the source languages, can naturally be assigned Gaussians with parameters $\langle \mu=0,\sigma=\sigma_{\mathit{av}}\rangle$ where $\sigma_{\mathit{av}}$ is the average variance in the estimated Gaussians. In our experiments, we rely on simple feature representations that are identical for all languages. (ii) Also, consider the obvious extension of using empirical Gaussian priors in the multi-source set-up, where we regularize the target to stay in one of several bounding ellipsoids rather than the one given by the full set of source models. These ellipsoids could come from typologically different groups of source languages or from individual source languages (and then have constant width). While this is technically a non-convex regularizer, we can simply run one model per source group and choose the one with the best fit to data. We do not explore this direction further in this paper. 

\subsection{Empirical Gaussian noise injection}

We also introduce a drop-out variant of empirical Gaussian priors. Our point of departure is average structured perceptron. 
We implement empirical Gaussian noise injection with Gaussians $\langle (\mu_1,\sigma_1),\ldots, (\mu_m,\sigma_m)\rangle$ for $m$ features as follows. We initialise our model parameters with the means $\mu_j$. For every instance we pass over, we draw a corruption vector $\mathbf{g}$ of random values $v_i$ from the corresponding Gaussians $(1,\sigma_i)$. We inject the noise in $\mathbf{g}$ by taking pairwise multiplications of $\mathbf{g}$ and our feature representations of the input sequence with the relevant label sequences. 
Note that this drop-out algorithm is parameter-free, but of course we could easily throw in a hyper-parameter controlling the degree of regularization. We give the algorithm in Algorithm 1. 

\begin{algorithm}
{\footnotesize
  \begin{algorithmic}[1]
    \STATE {$T=\{ \langle\mathbf{x}^1,\mathbf{y}^1\rangle,\ldots,\langle\mathbf{x}_n,\mathbf{y}_n\rangle\}\text{~w.~}\mathbf{x}_i=\langle v_1,\ldots\rangle\text{ and }v_k=\langle f_1,\ldots,f_m\rangle, \mathbf{w}^0=\langle w_1:\hat{\mu_1},\ldots, w_m:\hat{\mu_m}\rangle$}
   \FOR {$i\leq I\times |T|$}
   \FOR {$j\leq n$}
   \STATE $\mathbf{g}\leftarrow \mathbf{sample}(\mathcal{N}(1,\sigma_1),\ldots,\mathcal{N}(1,\sigma_m))$
   \STATE {$\hat{\mathbf{y}}\leftarrow \arg\max_{\mathbf{y}}\mathbf{w}^i \cdot \mathbf{g}$}
   \STATE {$\mathbf{w}^{i+1}\leftarrow\mathbf{w}^i+\Phi(\mathbf{x}_j,\mathbf{y}_j)\cdot \mathbf{g}-\Phi(\mathbf{x}_j,\hat{\mathbf{y}})\cdot \mathbf{g}$}
   \ENDFOR 
   \ENDFOR
  \end{algorithmic}
  \caption{\label{perc}Averaged structured perceptron with empirical Gaussian noise}
  }
\end{algorithm}

\section{Cross-lingual POS Experiments}
\paragraph{Data} In our multi-source cross-language transfer learning set-up, we rely on 15 multiple source language models to estimate our priors. We use a subset of the data in \citep{Agic:ea:15}. 

\paragraph{Annotation projection} We learn IBM-2 word alignment models from the Bible using EM to project annotation from the 15 source languages to our 10 target languages. We assign each word in each target language the majority vote tag after projecting from all source languages. 

\paragraph{Features} We use a simple feature template considering only orthographic features and cross-lingual word embeddings. The orthographic features include whether the current word contains capital letters, hyphens, or numbers. The embeddings are 40-dimensional distributional vectors capturing information about the distribution of words in a multi-parallel corpus. We learned these embeddings using an improvement over the technique suggested in \citet{Soegaard:ea:15}. \citet{Soegaard:ea:15} suggest a remarkably simple approach to learning distributional representations of words that transfer across languages. In parallel document collections or parallel corpora, we can represent the meaning of words by a vector encoding in what documents or sentences each word occurs. This is known as {\em inverted indexing}~in database theory. We encode the meaning of words by binary vectors encoding their presence in biblical verses and then apply SVD to reduce these vectors to 40 dimensions. Dimensionality was chosen for comparability with other publicly available bilingual embeddings. While this approach assumes fewer resources available, published results suggest that such representations are superior to previous work \citep{Soegaard:ea:15}. We improve on this approach by  shifting and row normalisation. We tuned the parameters on Danish development data. 

\paragraph{Baselines and systems} Our first baseline is L2-regularized CRF learned using L-BFGS. Our batch CRF systems are {\sc L2-Prior} and {\sc EmpGauss}.Our second baseline is an online averaged structured perceptron with L2 weight decay, learned using additive updates. We augment averaged structured perceptron with empirical Gaussian noise injection (Algorithm 2), leading to {\sc EmpGaussNoise}. 

\paragraph{Parameters}
In {\sc L2-Prior}, we set the variance to be the same for all parameters, namely equivalent to the regularization parameter in our L2-regularized baseline. The parameter was optimized on Danish development data. 

\paragraph{Results} We report the results of several systems: Our CRF models -- {\sc L2-CRF}, {\sc L2-Prior}, {\sc Multi-L2-Prior} and {\sc EmpGauss} -- as well as our online models -- {\sc L2-Perc} and {\sc EmpGaussNoise}. We present the macro-average performances across our 10 target languages below. We compute significance using Wilcoxon over datasets following \citet{Demsar:06} and mark $p<0.01$ by **. 

\begin{center}\begin{tabular}{ccc|cc}
\toprule
{\sc L2-CRF}&{\sc L2-Prior}&{\sc EmpGauss}&{\sc L2-Perc}&{\sc EmpGaussNoise}\\
\midrule
76.1&80.31$^{**}$&81.02$^{**}$&75.04&80.54$^{**}$\\
\bottomrule
\end{tabular}
\end{center}

\section{Observations}

We make the following additional observations: (i) Following the procedure in \citet{Zhu:ea:09:nips}, we can compute the Rademacher complexity of our models, i.e., their ability to learn noise in the labels (overfit). Sampling POS tags randomly from a uniform distribution, chance complexity is 0.083. With small sample sizes, L2-CRFs actually begin to learn patterns with Rademacher complexity rising to 0.086, whereas both {\sc L2-Prior} and {\sc EmpGauss} never learn a better fit than chance. (ii) \citet{Geman:ea:92} present a simple approach to explicitly studying bias-variance trade-offs during learning. They draw subsamples of $l< m$ training data points $\mathcal{D}_1, \ldots, \mathcal{D}_k$ and use a validation dataset of $m'$ data points to define the integrated variance of our methods. Again, we see that using empirical Gaussian priors lead to less integrated variance. (iii) An empirical Gaussian prior effectively limits us to hypotheses in $\mathcal{H}$ in a ellipsoid around the average source model. When inference is exact, and our loss function is convex, we learn the model with the smallest loss on the training data within this ellipsoid. Model interpolation of (some weighting of) the average source model and the unregularized target model can potentially result in the same model, but since model interpolation is limited to the hyperplane connecting the two models, the probability of this to happen is infinitely small ($\frac{1}{\infty}$). Since for any effective regularization parameter value (such that the regularized model is different from the unregularized model), the empirical Gaussian prior can be expected to have the same Rademacher complexity as model interpolation, we conclude that using empirical Gaussian priors is superior to model interpolation (and data concatenation).

\bibliographystyle{plainnat}
\bibliography{../../biblio}

\begin{thebibliography}{12}
\providecommand{\natexlab}[1]{#1}
\providecommand{\url}[1]{\texttt{#1}}
\expandafter\ifx\csname urlstyle\endcsname\relax
  \providecommand{\doi}[1]{doi: #1}\else
  \providecommand{\doi}{doi: \begingroup \urlstyle{rm}\Url}\fi

\bibitem[Agic et~al.(2015)Agic, Hovy, and S{\o}gaard]{Agic:ea:15}
Zeljko Agic, Dirk Hovy, and Anders S{\o}gaard.
\newblock If all you have is a bit of the {B}ible: Learning {POS} taggers for
  truly low-resource languages.
\newblock In \emph{ACL}, 2015.

\bibitem[Casella(1985)]{Casella:85}
George Casella.
\newblock An introduction to empirical {B}ayes data analysis.
\newblock \emph{American Statistician}, 39:\penalty0 83--87, 1985.

\bibitem[Collins(2002)]{Collins:02}
Michael Collins.
\newblock {Discriminative Training Methods for Hidden Markov Models: Theory and
  Experiments with Perceptron Algorithms}.
\newblock In \emph{EMNLP}, 2002.

\bibitem[Demsar(2006)]{Demsar:06}
Janez Demsar.
\newblock Statistical comparisons of classifiers over multiple data sets.
\newblock \emph{Journal of Machine Learning Research}, 7:\penalty0 1--30, 2006.

\bibitem[Geman et~al.(1992)Geman, Bienenstock, and Doursat]{Geman:ea:92}
Stuart Geman, Elie Bienenstock, and Rene Doursat.
\newblock Neural networks and the bias/variance dilemma.
\newblock \emph{Neural Computation}, 4:\penalty0 1--58, 1992.

\bibitem[Lafferty et~al.(2001)Lafferty, McCallum, and Pereira]{Lafferty:ea:01}
John Lafferty, Andrew McCallum, and Fernando Pereira.
\newblock {Conditional random fields: probabilistic models for segmenting and
  labeling sequence data}.
\newblock In \emph{ICML}, 2001.

\bibitem[Li et~al.(2012)Li, Gra\c{c}a, and Taskar]{Li:ea:12}
Shen Li, Jo{\~a}o Gra\c{c}a, and Ben Taskar.
\newblock Wiki-ly supervised part-of-speech tagging.
\newblock In \emph{EMNLP}, 2012.

\bibitem[Naseem et~al.(2010)Naseem, Chen, Barzilay, and Johnson]{Naseem:ea:10}
Tahira Naseem, Harr Chen, Regina Barzilay, and Mark Johnson.
\newblock {Using universal linguistic knowledge to guide grammar induction}.
\newblock In \emph{Proceedings of EMNLP}, 2010.

\bibitem[Smith and Osborne(2005)]{Smith:Osborne:05}
Andrew Smith and Miles Osborne.
\newblock Regularisation techniques for conditional random fields:
  Parameterised versus parameter-free.
\newblock In \emph{IJCNLP}, 2005.

\bibitem[S{\o}gaard(2013)]{Soegaard:13:naacl}
Anders S{\o}gaard.
\newblock Zipfian corruptions for robust pos tagging.
\newblock In \emph{Proceedings of NAACL}, 2013.

\bibitem[S\o{}gaard et~al.(2015)S\o{}gaard, Agi{\'c}, Alonso, Plank, Bohnet,
  and Johannsen]{Soegaard:ea:15}
Anders S\o{}gaard, {\v{Z}}eljko Agi{\'c}, H\'ector~Mart\'inez Alonso, Barbara
  Plank, Bernd Bohnet, and Anders Johannsen.
\newblock Inverted indexing for cross-lingual nlp.
\newblock In \emph{ACL}, 2015.

\bibitem[Zhu et~al.(2009)Zhu, Rogers, and Gibson]{Zhu:ea:09:nips}
Jerry Zhu, Timothy Rogers, and Bryan Gibson.
\newblock Human {R}ademacher complexity.
\newblock In \emph{NIPS}, 2009.

\end{thebibliography}

\end{document}